\documentclass[11pt,a4paper]{article}
\usepackage[hyperref]{AACL-IJCNLP2020}
\usepackage{times}
\usepackage{latexsym}

\usepackage{microtype}

\aclfinalcopy 
\def\aclpaperid{218} 

\usepackage{graphicx}
\usepackage{color}
\usepackage{tabularx}
\usepackage{array}
\usepackage{caption}
\usepackage{textcomp}
\usepackage{bm}
\usepackage{booktabs}
\usepackage{multicol} 
\usepackage{multirow}
\usepackage{bigstrut}
\usepackage{makecell}
\usepackage[utf8]{inputenc}
\usepackage{CJKutf8}
\usepackage{mathtools}
\usepackage{adjustbox}

\title{Lexically Cohesive Neural Machine Translation with Copy Mechanism}

\author{Vipul Mishra$^{\dag}$,~Chenhui Chu$^{\ddag}$,~Yuki Arase$^{\dag}$ \\
  $^{\dag}$~Graduete School of Information Science and Technology, Osaka University \\
  $^{\ddag}$~Graduete School of Informatics, Kyoto University \\
  $^{\dag}$~\texttt{\{mishra.vipul, arase\}@ist.osaka-u.ac.jp} \\
  $^{\ddag}$~\texttt{chu@i.kyoto-u.ac.jp} \\}

\date{}

\begin{document}
\maketitle
\begin{abstract}
Lexically cohesive translations preserve consistency in word choices in document-level translation. 
We employ a copy mechanism into a context-aware neural machine translation model to allow copying words from previous translation outputs. 
Different from previous context-aware neural machine translation models that handle all the discourse phenomena implicitly, our model explicitly addresses the lexical cohesion problem by boosting the probabilities to output words consistently.  
We conduct experiments on Japanese to English translation using an evaluation dataset for discourse translation. 
The results showed that the proposed model significantly improved lexical cohesion compared to previous context-aware models. 
\end{abstract}

\section{Introduction}
\label{intro}
As neural machine translation (NMT) significantly improved sentence-level translation qualities~\cite{google-human-parity,microsoft-human-parity}, recent studies have been focused on document-level translation. 
In particular, discourse in document-level translation is one of the central research interests, such as addressing coreference and anaphora resolution; and preserving cohesion and coherence in translation; {\em e.g.},~\cite{transformer-context-encoder,miculicich-etal-2018-han}. 
In this study, we tackle the problem of lexical cohesion, which aims to consistently use the same target words to translate the same source words. 
\newcite{lcohesion-metric} discussed that lexical cohesion significantly affects the overall quality of document translation.
Table~\ref{tab:inconsistent-translation} shows a comparison of lexically incohesive and cohesive translations for two consecutive Japanese sentences. 
The incohesive translations translate the same Japanese word ``\begin{CJK}{UTF8}{ipxm}時計\end{CJK}'' (watch) into ``clock'' and ``watch,'' while the cohesive translations consistently translate the word into ``watch.''

Previous studies approached discourse phenomena in NMT using a context-aware NMT model, which inputs previous source sentences and their translations as contexts. 
However,~\newcite{nagata-morishita-2020-test} showed that lexical cohesion is hard to solve with only context-aware models. 
We conjecture this is because context-aware models handle previous translations as a whole and are not sensitive enough to word usage consistency.

In this study, we employ a copy mechanism \cite{summarization-point-generate} on the context-aware NMT model for document-level translation~\cite{miculicich-etal-2018-han} to explicitly address the lexical cohesion problem. 
Our model computes a probability of copying a target word from previous translation outputs and boosts its output probability in the translation of a current sentence.

We conduct experiments on Japanese to English document translation.
The results indicate that our model achieves significantly better lexical cohesion, comparing to previous context-aware NMT models.

\begin{table*}[t]
\centering
\begin{adjustbox}{max width=\linewidth}
\def\arraystretch{1.2}
    \begin{tabular}{l|l}
    \hline
    \multirow{2}{*}{Source} & \begin{CJK}{UTF8}{ipxm}田中さん、よい\underline{時計}をお持ちですね。\end{CJK}\\
    & \begin{CJK}{UTF8}{ipxm}ありがとう、この\underline{時計}は祖父の形見なんです。\end{CJK} \\
    \hline
    \multirow{2}[2]{*}{Incohesive translations}& You have a good \underline{clock}, Mr. Tanaka.\\
    &Thank you, this \underline{watch} is a memento of my grandfather.\\
     \hline
    \multirow{2}[2]{*}{Cohesive translations} & You have a good \underline{watch}, Mr. Tanaka.\bigstrut[t]\\
    &Thank you, this \underline{watch} is a memento of my grandfather.\bigstrut[t]\\
    \hline
    \end{tabular}
    \end{adjustbox}
    \caption{Example of lexically incohesive and cohesive translation}
    \label{tab:inconsistent-translation}
\end{table*}

\section{Related Work}
Previous studies for document-level NMT proposed models that take preceding source sentences and their translations as contexts to translate the current sentence, which are called context-aware NMT. 
These contexts are expected to be useful for addressing discourse translation.

\subsection{Source-Side Context Models}
The initial context-aware NMT models took preceding source sentences as context.
The 2-to-2 model \cite{2-to-2} was the first example of context-aware NMT. 
\newcite{hierarchical-rnn} proposed to use hierarchical recurrent neural networks (RNNs) for modeling contextual information from the several of previous source sentences. 
\newcite{nyu-context-aware} added a separate encoder for encoding the context sentence. 
In contrast, \newcite{inter-sentence-gate} employed the same encoder for encoding the current source sentence and the previous source sentence, but used an inter-sentence gate to model the importance of each sentence. 
\newcite{anaphora-improvement} incorporated context information in both the encoder and the decoder using an multi-head attention mechanism~\cite{transformer-vaswani}. 

\subsection{Integrating Target-Side Context}
Following the success of incorporating source-side contexts, researchers proposed to consider also target-side contexts, {\em i.e.}, translation outputs of preceding sentences. \newcite{transformer-context-encoder} proposed to use a separate Transformer encoder for capturing discourse information on both sides. \newcite{discourse-evaluation} used a separate encoder for context sentences and studied three ways of concatenation, an attention gate, and hierarchical attention to combine the context to the current sentence. 
\newcite{voita-etal-2019-good} proposed a model with two multi-head attention sub-layers to incorporate discourse-related information from the source and target sides, respectively. 
\newcite{xiong-aaai-2019} proposed to use both side contexts to first generate translations for individual sentences and train a reward teacher to learn the ordering structure in a document, in order to produce discourse coherent translations. 
\newcite{memory-networks} proposed to use memory networks for modeling both the source and target side context. 
\newcite{continuous-cache} proposed a cache memory holding the decoder state and a context vector of the previous translated sentence, and used this cache for translating the current sentence. 
\newcite{dynamic-caching} proposed a dynamic cache and a topic cache for coherent translation. 
These cache based models are memory-intensive as they require additional memory for caching. 
\newcite{tan-etal-2019-hierarchical} proposed a hierarchical model with both a sentence encoder and a document decoder. 
Different from them, \newcite{miculicich-etal-2018-han} proposed the hierarchical attention networks for integrating both source and target context with a hierarchical architecture in a memory efficient manner. 
%

These previous studies confirmed the effectiveness of inputting previous translation outputs as contexts. 
These studies implicitly handle all the discourse phenomena; however, \newcite{nagata-morishita-2020-test} showed that a context-aware model better handles coreference resolution while struggles to achieve lexical cohesion in Japanese to English translation.  
We employ the copy mechanism to explicitly addresses the lexical cohesion problem, which was empirically confirmed as effective in our evaluation. 


\section{Preliminaries: Hierarchical Attention Networks}
\label{sec: preliminaries}
We employ the copy mechanism to the hierarchical attention networks (HAN) \cite{miculicich-etal-2018-han}, which is a Transformer~\cite{transformer-vaswani} based context-aware NMT model, for its strong performance in discourse translation. 
HAN is also memory-efficient compared to the cache based models. 
Nonetheless, our model can be easily applied to other Transformer based context-aware NMT models.


\begin{figure*}[t]
\begin{center}
    \hspace*{\fill}
      \includegraphics[scale=0.55]{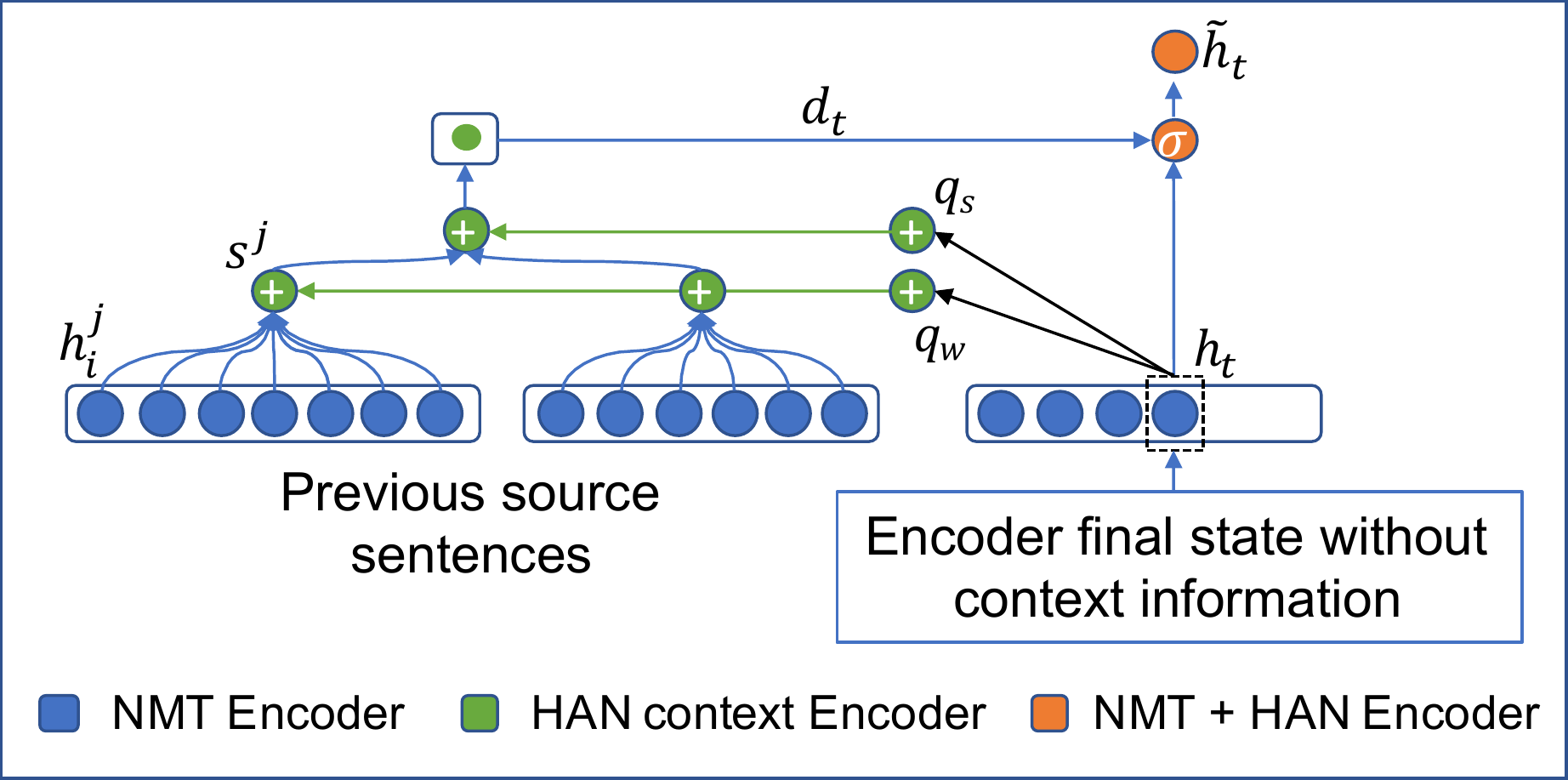}
    \hspace*{\fill}
    \caption{The encoder in the hierarchical attention networks}
    \label{fig:HAN-enc}
    \end{center}
\end{figure*}

The HAN model generates translation $\bm{y}$ for source sentence $\bm{x}$ by using information not only from $\bm{x}$ but also from the previous $n$ source sentences and their translation outputs. 
There are three settings for the HAN model depending on what contextual information is used in which part of the model. 
The HAN encoder model only uses the source-side context in the encoder. 
Similarly, the HAN decoder model only uses the target-side context, {\em i.e.}, previous translation outputs, in the decoder.
The HAN joint model is a combination of the HAN encoder model and HAN decoder model, where both the source-side and target-side contexts are used in the encoder and decoder, respectively.
Figure \ref{fig:HAN-enc} shows the process of integration of contextual information at time step $t$ in the HAN encoder model (similar process is conducted in the HAN decoder model).

Let us denote the set of $n$ preceding source sentences used as context by $\bm{D}_{x}=\{\bm{x}^1,\cdots,\bm{x}^n\}$ and their translations used as target-side context by $\bm{D}_{y}=\{\bm{y}^1,\cdots,\bm{y}^n\}$. 
The multi-head attention is used for representing relevant context information from the entire context in two steps. 
The multi-head attention takes a query, key, and value as inputs, and linearly projects them for $m$ times. 
On each of these projected queries, keys, and values, an attention function is computed in parallel: 
\begin{equation}
    \textrm {Attention}(Q, K, V)=\textrm{softmax}\left(\frac{Q K^{T}}{\sqrt{r}}\right) V,\nonumber
\label{eqn:scaled-dot}
\end{equation}
where $Q$, $K$, and $V$ are a projected query, key, and value, respectively,\footnote{In practice, multi-head attention is computed on a set of queries simultaneously, hence the inputs are matrices packing queries, keys, and values, respectively.} and $r$ is the dimension of the query and key. 
The $\textrm{softmax}(\cdot)$ is a softmax function. 
The outputs of $m$ attention heads concatenated and once again projected, resulting in the final output of the multi-head attention function.

In the first step of HAN, the word-level information of $j$-th sentence is summarized into a vector $\bm{s}^{j}$ as a sentence-level context:
\begin{eqnarray}
    \bm{s}^j &=&  \underset{i}{\textrm{MultiAtt}}\left(\bm{q}_{w}, \bm{h}_{i}^{j}, \bm{h}_{i}^{j}\right),\label{eqn:word_level_attn}\\
    \bm{q}_{w} &=& f\left(\bm{h}_{t}\right),\nonumber
\end{eqnarray}
where $\textrm{MultiAtt}(\cdot)$ represents multi-head attention, and $\bm{q}_{w}$ and $\bm{h}_{i}^{j}$ correspond to query, key, and value, respectively.
$\bm{q}_{w}$ is computed using $\bm{h}_{t}$ by linear transformation denoted as a function $f$, where $\bm{h}_{t}$ is the hidden state of the word to be encoded (or decoded) at time step $t$.  
$\bm{h}_{i}^{j}$ is the hidden state of the $i$-th word in the $j$-th context sentence. 
In the second step, the sentence-level context vectors of all $n$ sentences are transformed into a context vector $\bm{d}_{t}$ as a document-level context:
\begin{eqnarray}
    \bm{d}_{t}&=&\textrm{FFN}\left (\underset{j}{\textrm{MultiAtt}}\left(\bm{q}_{s}, \bm{s}^{j}, \bm{s}^{j}\right)\right),
    \label{eqn:sent-level-attn}\\
    \bm{q}_{s}&=&g\left(\bm{h}_{t}\right),\nonumber
\end{eqnarray}
where FFN is a position-wise feed-forward network, and $g$ is a linear transformation function.

The document-level context vector $\bm{d}_{t}$ is then integrated with the hidden state $\bm{h}_t$ of the current time step $t$. 
For integration, HAN uses a gating mechanism as: 
\begin{eqnarray}
    \widetilde{\bm{h}}_{t}=\lambda_{t} \bm{h}_{t}+\left(1-\lambda_{t}\right) \bm{d}_{t},\nonumber\\
    \lambda_{t}=\sigma\left(W_{h} \bm{h}_{t}+W_{d} \bm{d}_{t}\right),\nonumber 
\end{eqnarray}
where $W_{h}$ and $W_{d}$ are learnable parameter matrices and $\sigma(\cdot)$ is a sigmoid function.

In the case of the decoder, the final hidden state $\widetilde{\bm{h}}_{t}$ is passed through a softmax layer, which gives a probability distribution $\bm{P}_{\rm{\rm{vocab}}}$ over the target vocabulary: 
\begin{eqnarray*}
\bm{P}_{\rm{vocab}}=\textrm{softmax}(\widetilde{\bm{h}}_{t}).
\end{eqnarray*}
The word with the highest probability is generated. 

\begin{figure*}[t]
\begin{center}
      \includegraphics[scale=0.45]{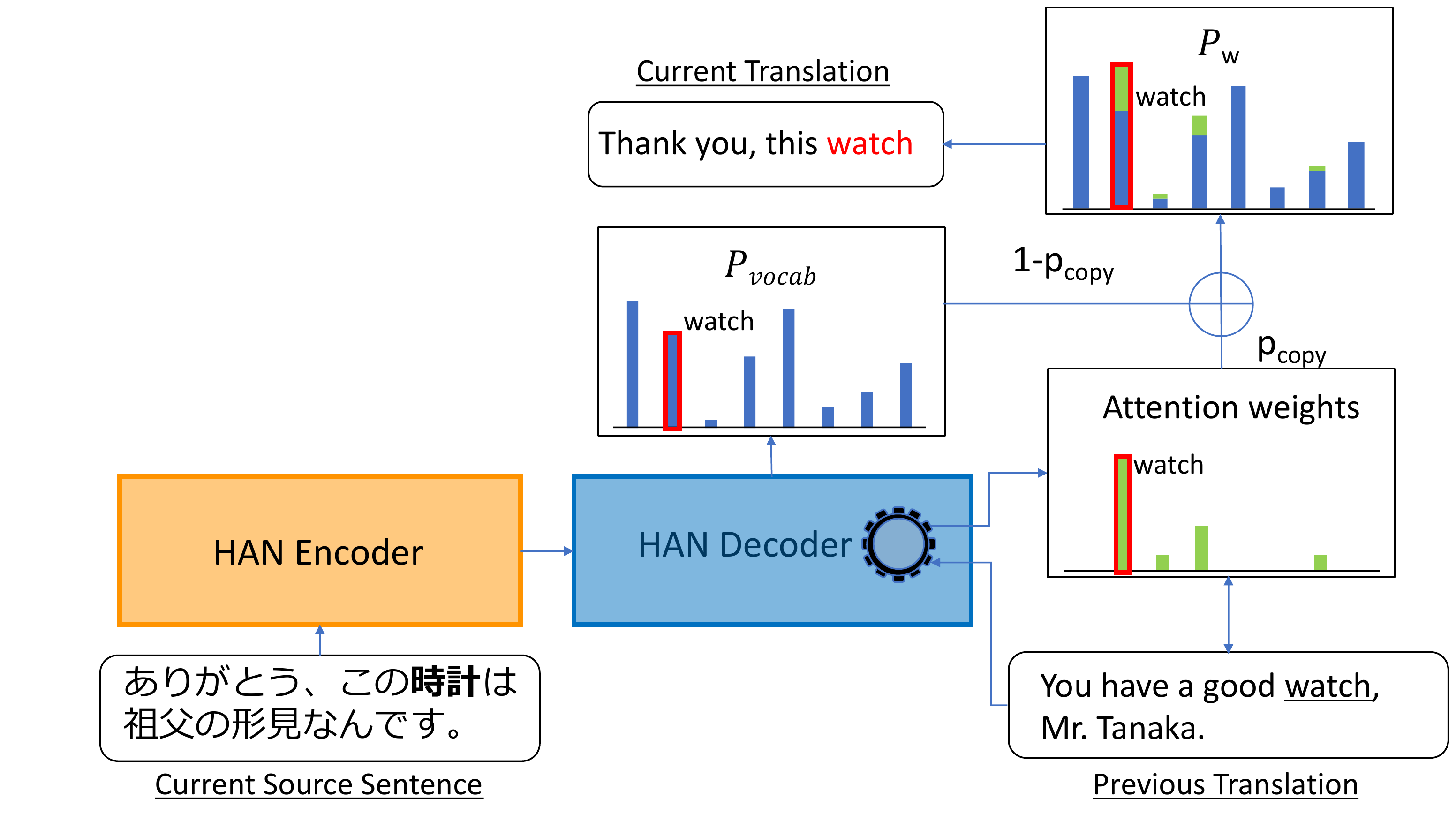}
\caption{Copy mechanism boosts the probability of outputting a cohesive word used in previous translations. It computes a copy probability $p_{\rm{copy}}$ and weight of a word using attention weights.}
\label{fig:proposed model}
\end{center}
\end{figure*}

\section{Integration of Copy Mechanism to Hierarchical Attention Networks}
\label{sec:proposed}
We employ a copy mechanism to \textit{\rm{copy}} target words from the previous translations to achieve lexical cohesion.  
In this study, we integrate the copy mechanism to the decoder of the HAN joint model, so that our model can benefit from the context representations of both preceding source sentences and their translations. 
The encoder of our model is exactly the same as that of the HAN joint model. 
The decoder is enhanced with the copy mechanism, which computes a probability to copy words from previous translations $\bm{D}_{y}$ and adjust $\bm{P}_{\rm{vocab}}$ to boost the probability to output these words. 
Figure \ref{fig:proposed model} shows how our model produces a cohesive translation for the second source sentence in Table \ref{tab:inconsistent-translation}. 
The final probability of outputting ``watch'' is boosted by a copy probability and its weight estimated using an attention mechanism.

We first calculate the copy probability $p_{\rm{copy}}$, which acts as a gate function to decide how much emphasis should be given to $\bm{P}_{\rm{vocab}}$ computed based on intra-sentential information and to probabilities for outputting words used in $\bm{D}_{y}$. 
The higher the value of $p_{\rm{copy}}$ becomes, the more likely it is that a word from $\bm{D}_{y}$ is to be copied and output at time step $t$. 
$p_{\rm{copy}}$ is calculated using the decoder hidden state $\widetilde{\bm{h}}_{t}$,  attentions to the encoder $\bm{c}_{t}$, and document-level context vector $\bm{d}_{t}$. 
\begin{equation}
\label{eq:p_copy}
p_{\rm{copy}}=\sigma\left(W_{\widetilde{h}}\widetilde{\bm{h}}_{t}+W_{c}\bm{c}_{t}+W_{dy}\bm{d}_{t}+\bm{b}\right),
\end{equation}
\noindent where $W_{\widetilde{h}}$, $W_{c}$, $W_{dy}$, and $\bm{b}$ are learnable parameters. 
$\bm{c}_{t}$ is computed through multi-head attention: 
\begin{eqnarray*}
    \bm{c}_t = \textrm{MultiAtt}\left(\widetilde{\bm{h}}_{t}, \widetilde{\bm{h}}_{enc}, \widetilde{\bm{h}}_{enc} \right),
\end{eqnarray*}
where the decoder hidden state $\bm{\widetilde{h}}_{t}$ is the query and the encoder hidden state $\widetilde{\bm{h}}_{enc}$ is the key and the value.
Note that $\widetilde{\bm{h}}_{enc}$ represents all the encoder hidden states of all the words in the current sentence. 

\begin{table*}[h!]
    \begin{center}
    \begin{tabular}{ l|cccccc}
    \hline
    Data split & \# sentences & \# src tokens & \# tgt tokens & src-length & tgt-length & \# talks\\
    \hline
    Train       & $223$k & $4.8$M   & $4.6$M  & $21.5$ & $20.4$ & $1,863$\\ 
    Validation  & $1,285$ & $27.7$k  & $24.7$k & $21.5$ & $19.2$ & $15$\\
    Test        & $1,194$ & $29.2$k  & $23.7$k & $24.8$ & $19.9$ & $12$\\
    \hline
    \end{tabular}
    \caption{Statistics of IWSLT 2017 Japanese-English task dataset}
    \label{tab:data_statictics}
    \end{center}
\end{table*}

Next, we compute weights of target words used in previous translations to copy at time step $t$. 
Attention weights pointing to the previous translations imply the importance of each previously translated target word when generating the current target word, and thus we use them as the weights for copying. 
We obtain an attention weight $a_{j}$ for $j$-th sentence $\bm{y}^j \in \bm{D}_{y}$ and an attention weight $a_{j,i}$ for each word $w_i \in \bm{y}^j$ from Equations~(\ref{eqn:word_level_attn}) and~(\ref{eqn:sent-level-attn}).  
When we use multi-head attention of $m$-heads, there are $m$ sets of $a_{j}$ and $a_{j,i}$, hence, we take the average attention weights for all heads.

The final attention weight $\alpha_i$ for $w_i$ at time step $t$ is:
\begin{eqnarray*}
    \alpha_{i}=\frac{1}{m^2} \sum_{\bm{y}^j \in \bm{D}_{y}}\sum_{\ell=1}^{m} a_{j}^{\ell}\sum_{\ell=1}^{m}a_{j,i}^{\ell},
    \label{eqn:prop_attn_calc}
\end{eqnarray*}
where superscript ${\ell}$ represents the ${\ell}$-th attention head.
Note that the value of $\alpha$ for words that did not appear in previous translations is zero.

Finally, the probability distribution to output target vocabulary $\bm{P}_{w}$ at time step $t$ is computed as:
\begin{equation}
    \bm{P}_{w} = \left(1 - p_{\rm{copy}}\right) \bm{P}_{\rm{vocab}} + p_{\rm{copy}}\bm{\alpha},
    \label{eqn:modify-distrib}
\end{equation}
where $\bm{\alpha}$ is a vector of attention weights for all words in the target vocabulary.
In Figure \ref{fig:proposed model}, the final probability of generation of ``watch'' becomes highest in the target vocabulary, boosted by the copy probability and the attention weight computed from Equation~(\ref{eqn:modify-distrib}).

\section{Experimental Settings}
To evaluate effects of the proposed model to generate lexically cohesive translation, we conducted experiments on Japanese to English translation.\footnote{Because there is no English to Japanese document translation test set is unavailable, we evaluated only Japanese to English direction.} 

\subsection{Dataset}
Following \newcite{miculicich-etal-2018-han} who conducted experiments on TED talk data for document translation, we used the IWSLT 2017 Ted talk Japanese-English task dataset \cite{iwslt-overview} for our experiments. 
Test2014 and test2015 were used for validation and testing, respectively. The dataset comprises of spoken language data, particularly, subtitle data from TED talks and their corresponding Japanese translations. The statistics of these datasets are presented in Table \ref{tab:data_statictics}.


To ensure that the copy mechanism performs word-level copying, we used words as a translation unit. 
Japanese sentences were segmented with Mecab\footnote{\url{https://taku910.github.io/mecab/}} and English sentences were tokenized with the tokenizer from the Moses toolkit.\footnote{\url{https://github.com/moses-smt/mosesdecoder/tree/master/scripts/tokenizer}}


\subsection{Evaluation Metrics}
BLEU~\cite{papineni-bleu} is commonly used to evaluate the overall translation quality; however, it does not have the power to test lexical cohesion. 
Hence, we also used two metrics proposed by previous studies.

\begin{description}
\item[BLEU]
We evaluated BLEU-4 scores on the IWSLT test2015 test set.

\item[Lexical Cohesion (LC)]
The LC score was proposed by \newcite{lcohesion-metric}, which has also been used in \cite{miculicich-etal-2018-han}. 
The LC score is defined as the percentage of the number of repetitively used words and lexically similar words to the number of content words presented in document-level translations. 
A lower LC score indicates that translations are lexically incohesive, while too high LC score means that translations contain over usage of the same target words. The latter is known as the over-generation problem, which is commonly found in NMT outputs. 
Due to this nature, good translations should result in an LC score closer to that of human references. 
In our evaluation, the LC score was again calculated on the IWSLT test2015 test set at document-level.

\end{description}


\begin{table*}[t]
    \begin{center}
    \begin{tabular}{ l | c | c | c c}
    \hline
    Model & \# context sentences &  BLEU  & LC \\ 
    \hline
    Sentence-level biRNN & $0$ & $13.4$ & $54.1$ ($-0.8$) \\ 
    Sentence-level Transformer & $0$ & $13.3$ & $48.6$ ($-6.3$) \\ \hline
    2-to-2 biRNN  & $1$ & $12.4$ & $53.3$ ($-1.6$) \\
    2-to-2 Transformer & $1$ & $12.8$ & $52.6$ ($-2.3$) \\
    HAN encoder & $1$ & $14.6$ & $55.6$ ($+0.7$) \\
    HAN decoder & $1$ & $14.6$ & $56.4$ ($+1.5$) \\
    HAN joint & $1$ & $14.7$ & $57.3$ ($+2.4$) \\ \hline
    \multirow{2}{*}{Proposed model} & $1$ & $\bm{14.9}$ & $55.3$ ($+0.4$) \\ 
      & $3$ & $14.5$ & $\bm{55.1}$ ($\bm{+0.2}$) \\ \hline 
    Human reference & -- & -- & $54.9$ \\
    \hline
    \end{tabular}
    \caption{Translation quality and lexical cohesion scores. BLEU and LC scores were measured on IWSLT test2015. Numbers in parentheses for the LC scores denote the difference between the human reference.}
    \label{tab:final_result}
    \end{center}
\end{table*}

\subsection{Comparison}
We evaluated our model in Japanese to English translation comparing to the following strong baselines with different network architectures. 

\paragraph{Sentence-level NMT models:}
    \begin{itemize}
        \item Sentence-level biRNN: the bidirectional RNN based NMT model with attention, proposed by \newcite{DBLP:journals/corr/BahdanauCB14}.
        \item Sentence-level Transformer: the Transformer model of \newcite{transformer-vaswani}.
    \end{itemize}
\paragraph{Context-aware NMT models:}
    \begin{itemize}
        \item 2-to-2 biRNN: A simple yet effective approach that concatenates the current sentence with the previous sentence on both source and target sides, and input to the model \cite{2-to-2}. Here the base architecture is the biRNN attention based NMT model.
        \item 2-to-2 Transformer: A 2-to-2 approach with Transformer.
        \item HAN encoder: the HAN model integrating information only from previous source sentences at the encoder.
        \item HAN decoder: the HAN model only integrating information only from previous translation outputs at the decoder.
        \item HAN joint: the HAN model integrating both preceding source sentences and their translation outputs as contexts in the encoder and decoder, respectively.
    \end{itemize}

\subsection{Implementation and Training Details}
We used OpenNMT-py\footnote{\url{https://github.com/OpenNMT/OpenNMT-py/}} \cite{opennmt} to implement all the models compared in our experiments. 

\paragraph{Sentence-level NMT Models}
\label{subsec:openNMT-config}
For both the biRNN and Transformer sentence-level models, we used the recommended settings in OpenNMT-py. 
The dimensions for the source and target embeddings were set to $512$. 
The biRNN has $2$ layers with hidden size of $500$. 
The Transformer has $6$ layers and $8$ heads for multi-head attention, where dropout rate of $0.1$ was applied to Transformer blocks. 
The source and target vocabularies were set to $50$k. 
The mini-batch size was set to $4,096$ tokens. 
We used the Adam optimizer \cite{DBLP:journals/corr/KingmaB14} with default settings to optimize parameters of each model.
We trained all models for $150$k steps and tested ones with the lowest validation loss. 

\paragraph{Context-Aware NMT Models}
The settings for both of the 2-to-2 biRNN and Transformer were the same as those of the sentence-level biRNN and Transformer, respectively, besides that the maximum sentence length was set to $100$.

For the HAN models, we used the implementation provided by the authors.\footnote{\url{https://github.com/idiap/HAN_NMT}} 
Following \newcite{miculicich-etal-2018-han}, we first trained a sentence-level Transformer model for $20$ epochs using the recommended settings in OpenNMT-py. 
After that, the model with the lowest validation loss was fine-tuned for $1$ epoch to incorporate document-level context in the HAN models. 
The mini-batch size was set to $1,024$ tokens.

For the HAN encoder model, the number of previous source sentences to consider as context was set to $1$. 
The model was fine-tuned for $1$ epoch to tune the parameters for integrating the encoder context. 
The HAN decoder model was trained in the same way as the HAN encoder model but used $1$ previously translated target sentence as context at the decoder. 
During fine-tuning, the parameters for integrating the decoder context was updated. 
The HAN joint model is the combination of HAN encoder and decoder models. 
It used the HAN encoder model as the pre-trained model and then fine-tuned the decoder for $1$ additional epoch using the target context of $1$ previously translated target sentence. 

\begin{table*}[t]
    \centering
\begin{adjustbox}{max width=\linewidth}
\def\arraystretch{1.2}
    \begin{tabular}{c|p{0.9\linewidth}}
    \hline
    \multirow{2}{*}{Source} & \begin{CJK}{UTF8}{ipxm}\underline{ハト}と友達の人なら誰でも言うように、手の上で食事をするようにその鳥を訓練するのはそこまで努力はかからない。\end{CJK}\\
    &\begin{CJK}{UTF8}{ipxm}実際、\underline{ハト}は中に鳥の餌があると知っているのなら、喜んで玄関の中に歩いてくるでしょう。\end{CJK}\\
    \hline
     \multirow{2}{*}{Ref.} & as anyone who has befriended a \underline{pigeon} will tell you, it doesn't take much effort to train the bird to eat out of your hand.\\
     &In fact, a \underline{pigeon} will happily walk through your front door if it knows there is birdseed inside.\\
    \hline
    \multirow{2}{*}{HAN}& if you're a \underline{dove} and a friend, you can say that you can train a bird to eat on your hand, it's not that hard.\\
     &in fact, if you know that the \underline{pigeon} is feeding, you will be happy to walk in the door.\\
     \hline
     \multirow{2}{*}{Proposed}& if you're a \underline{pigeon} and a friend of mine, it's not that hard to train that bird to eat on your hand.\\
     &in fact, if you know that the \underline{pigeon} is feeding, you will be happy to walk in the door.\\
    \hline \hline
    
     \multirow{2}{*}{Source }&\begin{CJK}{UTF8}{ipxm}しかし虫や微生物や粒子レベルになると ファイの\underline{量}は低下します。\end{CJK}\\
    &\begin{CJK}{UTF8}{ipxm}情報統合の\underline{量}が低下しても ゼロにはなりません。\end{CJK}\\
    \hline
    \multirow{2}{*}{Ref.} &but as you go down to worms, microbes, particles, the \underline{amount} of phi falls off.\\
     &the \underline{amount} of information integration falls off, but it's still non-zero.\\
    \hline
    \multirow{2}{*}{HAN}& but when you get to insects and microbes at a level, you're going to reduce the \underline{amount} of moisture you get.\\
    &it's not zero information integration.\\
    \hline
    \multirow{2}{*}{Proposed}& but when it comes to insects and microbes and particles, the \underline{amount} of them go down.\\
     &it doesn't mean that the \underline{amount} of information integration goes down.\\
    \hline \hline
    
    \multirow{2}{*}{Source} & \begin{CJK}{UTF8}{ipxm}\underline{情報処理}が行われるところにはどこでも \underline{意識}があるのです。\end{CJK}\\
    &\begin{CJK}{UTF8}{ipxm}人間が行うような複雑な\underline{情報処理}には複雑な\underline{意識}が伴います。\end{CJK}\\
    \hline
    \multirow{2}{*}{Ref.} & wherever there's \underline{information processing}, there's \underline{consciousness}.\\
     &complex \underline{information processing}, like in a human, complex \underline{consciousness}.\\
    \hline
    \multirow{2}{*}{HAN} & there's a sense everywhere where \underline{information} is available.\\
    &there's complex \underline{consciousness} in complex ways that people do it.\\
    \hline
    \multirow{2}{*}{Proposed}& everywhere I get \underline{information processing}, I have \underline{consciousness} everywhere.\\
     &complex \underline{information processing} that humans do is complex.\\
    \hline
    \end{tabular}
    \end{adjustbox}
    \caption{Translation outputs of the HAN joint and proposed models.
    }
    \label{tab:examples-of-translation}
\end{table*}

\paragraph{Proposed Model}
We trained our model in the same manner as the HAN joint model.
We used the HAN encoder model as a pre-trained model and fine-tuned the parameters for target-side contextual information and also parameters for the calculation of $p_{\rm{copy}}$ (Equation~(\ref{eq:p_copy})) for $1$ epoch. 

We also investigated the influence of the number of context sentences in our model.
\begin{description}
    \item[$1$ context sentence:] used only $1$ previous source sentence and its translation as contexts.
    \item[$3$ context sentences:] used $3$ previous source sentences and their translations as contexts. 
\end{description}

\section{Results and Discussion}
Table \ref{tab:final_result} shows the results. 
The 2-to-2 models decreased the BLEU scores compared to the sentence-level conterparts. 
This is because 2-to-2 models simply concatenates a previous source sentence to input, making the sequence length doubled. 
NMT models are sensitive to sequence length and longer sequences are harder to generate \cite{DBLP:journals/corr/BahdanauCB14,transformer-vaswani}. 
In contrast, the HAN and proposed models outperformed sentence-level models. 
They show comparable BLEU scores; both use context representations that were effective to improve the BLEU scores compared to sentence-level models. 
Increasing the number of context sentences in the proposed model only slightly affect BLEU scores, indicating that it is robust to the size of context in terms of translation quality.

As for LC scores, the closer they are to that of the human references, the better lexical cohesion are achieved.
Our model had the most close LC score to the human references among all the models. 
The longer contexts were shown to be effective in our model. 
As expected, sentence-level models lack lexical cohesion. 
2-to-2 models was ineffective to improve lexical cohesion. 
The HAN models resulted in scores largely higher than that of the human references. 
This is caused by over-generation of personal pronouns, which has also been pointed by \newcite{miculicich-etal-2018-han}.


Table \ref{tab:examples-of-translation} shows three translation outputs of our model compared to these of the HAN joint model and human references.
In the first example, the HAN joint model output translations lexically incohesive; the Japanese word \begin{CJK}{UTF8}{ipxm}``ハト''\end{CJK} was inconsistently translated into ``dove'' and ``pigeon'' in two consecutive sentences, respectively. 
Our model consistently translated the word into ``pigeon'' in two sentences thanks to the copy mechanism.
In the second example, the proposed model consistently translated the Japanese word \begin{CJK}{UTF8}{ipxm}``量''\end{CJK} into ``amount.'' 
In contrast, the HAN joint model failed to generate, {\em i.e.}, under-generated the ``amount'' in the second translation. 
As a side effect of the copy mechanism, the proposed model incorrectly copied the word ``down'' at the end, changing the meaning of the source sentence. 
In the third example, our model consistently translated \begin{CJK}{UTF8}{ipxm}``情報処理''\end{CJK} to ``information processing,'' while the HAN joint model failed to complete translation of these words. 

These examples show that the proposed model earned the ability to properly copy target words from previous translations. 
Interestingly, not only lexical cohesiveness but also the overall translation quality was also improved in some cases. 
This may be because copied words have a kind of word sense disambiguation effects in the decoder. 
Further investigation is our future work. 
As a side effect of our model, sometimes it failed to choose a correct word to copy. 
We conjecture this is because we rely only on attention weights to compute weights of words to copy. 
We leave this problem for our future work.


\section{Conclusion}
We employed a copy mechanism to address the lexical cohesion problem in document-level NMT. 
Our model computes a copy probability and weights of words to copy referring to preceding source sentences and their translation outputs. 
Experiments on Japanese to English translation indicated that our model is effective to improve lexical cohesion, compared to strong context-aware NMT models. 
As future work, we intend to evaluate the effectiveness of our model on various language pairs and domains, such as English-French and English-Russian; news and novels. 
Also, we will improve the weighting method to copy words to avoid copying inappropriate words. 


\bibliography{aacl-ijcnlp2020}
\bibliographystyle{acl_natbib}

\end{document}